\documentclass{article}     
\usepackage{bbm}
\usepackage{times}
\usepackage{soul}
\usepackage{url}
\usepackage[hidelinks]{hyperref}
\usepackage[utf8]{inputenc}
\usepackage[small]{caption}
\usepackage{graphicx}
\usepackage{amsmath}
\usepackage{amsthm}
\usepackage{booktabs}
\usepackage{algorithm}
\usepackage{algorithmic}
\usepackage[algo2e]{algorithm2e}
\usepackage{verbatim}
\usepackage{graphicx}
\usepackage{subfigure}
\usepackage{amsmath}
\usepackage{bm}
\usepackage{booktabs}
\usepackage{color}
\usepackage{amsmath,epsfig,bm, amssymb,subfigure,graphicx,algorithm,algorithmic,color}
\usepackage{color}

\newcommand{\argmin}{\mathop{\mathrm{argmin\,}}}

\newcommand{\boldone}{{\boldsymbol{1}}}

\newcommand{\boldA}{{\boldsymbol{A}}}

\newcommand{\boldH}{{\boldsymbol{H}}}
\newcommand{\boldI}{{\boldsymbol{I}}}

\newcommand{\boldK}{{\boldsymbol{K}}}
\newcommand{\boldL}{{\boldsymbol{L}}}

\newcommand{\boldW}{{\boldsymbol{W}}}
\newcommand{\boldX}{{\boldsymbol{X}}}

\newcommand{\boldZ}{{\boldsymbol{Z}}}

\newcommand{\boldf}{{\boldsymbol{f}}}

\newcommand{\boldx}{{\boldsymbol{x}}}
\newcommand{\boldy}{{\boldsymbol{y}}}

\newcommand{\boldbeta}{{\boldsymbol{\beta}}}

\newcommand{\calS}{{\mathcal{S}}}

\advance\oddsidemargin-1in
\title{GraphLIME:Local Interpretable Model Explanations \\for Graph Neural Networks}

\author{
Qiang~Huang$^1$,%\footnote{Work done at RIKEN AIP},
Makoto~Yamada$^{2,3}$,
Yuan Tian$^1$,
Dinesh Singh$^3$,
Yi Chang$^1$\\
$^1$Jilin University, $^2$Kyoto University, $^3$RIKEN AIP,\\
huangqiang18@mails.jlu.edu.cn, myamada@i.kyoto-u.ac.jp, yuantian@jlu.edu.cn\\ dinesh.singh@riken.jp, yichang@jlu.edu.cn
}

\begin{document}

%\onecolumn
\maketitle
%\onecolumn

\begin{abstract}
 Graph structured data has wide applicability in various domains such as physics, chemistry, biology, computer vision, and social networks, to name a few. Recently, graph neural networks (GNN) were shown to be successful in effectively representing graph structured data because of their good performance and generalization ability. GNN is a deep learning based method that learns a node representation by combining specific nodes and the structural/topological information of a graph. However, like other deep models, explaining the effectiveness of GNN models is a challenging task because of the complex nonlinear transformations made over the iterations. In this paper, we propose GraphLIME, a local interpretable model explanation for graphs using the Hilbert-Schmidt Independence Criterion (HSIC) Lasso, which is a nonlinear feature selection method. GraphLIME is a generic GNN-model explanation framework that learns a nonlinear interpretable model locally in the subgraph of the node being explained. More specifically, to explain a node, we generate a nonlinear interpretable model from its $N$-hop neighborhood and then compute the $K$ most representative features as the explanations of its prediction using HSIC Lasso. Through experiments on two real-world datasets, the explanations of GraphLIME are found to be of extraordinary degree and more descriptive in comparison to the existing explanation methods.

% In recent years, because of the powerful expression of graph structure, the research of graph analysis by machine learning method has been paid more attention. Graph Neural Network (GNN) is a deep learning based method to process graph domain information. GNN has recently become a widely used graph analysis method due to its good performance and powerful generalization ability on graphs. However, explaining predictions made by GNNs is still a challenging task because GNNs combine both feature and graph structure information of nodes, which leads to a complex and nonlinear deep model. In this paper, we propose Local Interpretable GNN-Model Explanations Based on HSIC Lasso (GraphLIME), which is a general GNN-model explanation framework by learning a nonlinear interpretable model locally in the subgraph of a node being explained. More specially, we sample the neighboring nodes of the node being explained in $N$-hop network, and generate a nonlinear interpretable model which can find most representative features with Hilbert-Schmidt Independence Criterion Lasso (HSIC Lasso), finally we select top-$K$ features as the explanations of the prediction based on the nonlinear model. Through experiments on two real world graph datasets, we show that the proposed method compares favorably with the existing explanation methods.
\end{abstract}

\section{Introduction}

{D}{eep} Neural Network (DNN) is essentially a new machine learning algorithm based on discriminant model. DNN can model for complex and nonlinear problems and learn the underlying features of data to obtain more abstract features, which can improve the model's capability for prediction or classification. It has a large number of applications such as image recognition, automatic speech recognition, disease diagnosis, etc \cite{liu2017survey}%lecun2015deep,noda2015audio}.

Although the high degree of nonlinearity gives DNNs a powerful model representation capability, DNN is a black-box model, which make the model and its predictions hard to be interpreted. Also, people usually will not use a model or a prediction if they do not trust it. For example, if there is a medical diagnosis system that has high accuracy but can not present faithful explanations for its decisions, doctors will not use it. Therefore, it is very important to explore what a DNN model learns from data and why it takes a particular decision in a way that humans can understand. Recent works aimed at interpreting general DNNs mainly focus on two research routes. One approach is to locally approximate models with a simple and interpretable model such as linear regression, which can itself be probed for explanations \cite{lakkaraju2017interpretable,ribeiro2016should}. The other one is to examine models for relevant components such as identifying the most representative features in the input data \cite{chen2018learning,lundberg2017unified,sundararajan2017axiomatic} or influential input instances \cite{koh2017understanding,yeh2018representer}.

%\item Recently, the GNN is pretty popular and there are many applications. Actually, there exist a large number of interpretation method in DNN. But, there is a very few method to interpret GNN
\begin{figure}[t]
%\hspace*{-1.2em}
\centering
\centering
\includegraphics[width=12cm]{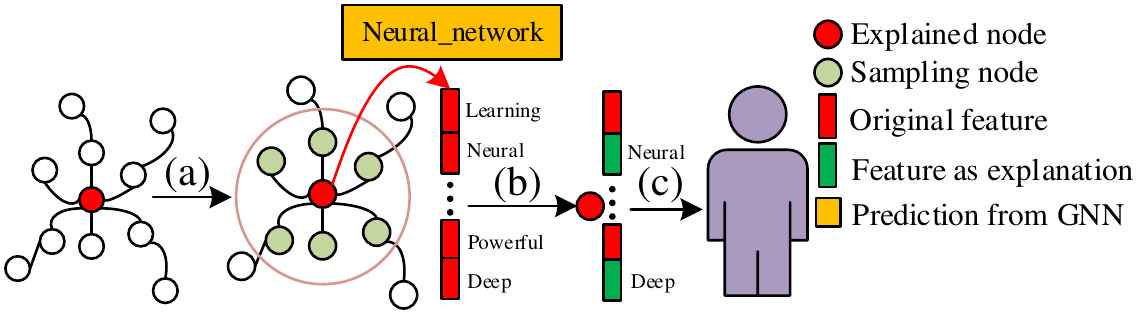}
%\caption{fig2}
\centering
\caption{Sample neighbors of the explained red node; (b) Obtain finite numbers of features as explanations (the green part of the features of the red node); (c) Present explanations to users.\label{GraphLIME}}
\end{figure}
In many real-world cases, there are considerable amounts of data without regular spatial structures, called non-Euclidean data, and they can be naturally represented as a graph, for example, graph data extracted from social networks, citation networks, electronic transactions, protein structures, molecular structures, and so on. Thus, modeling and analyzing graph data is a challenging task since it needs to combine both feature information of nodes as well as graph information together.
Currently, Graph Neural Networks (GNNs) are widely used because of their powerful modeling capability with regard to graphs. GNNs use neural networks to incorporate the feature information of nodes in a graph and as well as the structure information and pass these messages through the edges of the graph in non-Euclidean domains. However, GNNs are notoriously difficult to interpret, and their predictions are hard to explain, similar to that in DNNs. It is important to develop an interpretation method for GNNs because it can improve the transparency of a GNN model and contribute to getting humans to trust the model. Although there exists a large number of interpretation methods designed for DNNs, these methods are not suited for GNN-model interpretation because they do not explicitly use the graph information but perform in Euclidean domains. 

Recently, GNNexplainer \cite{ying2019gnnexplainer} was proposed. It can find the subgraph and select features of the explained node as explanations, but it mainly focuses on graph structures and not on finding useful features. An alternative idea is to use LIME \cite{ribeiro2016should}, which uses a linear explanation model to find features as explanations for GNN. However, the performance of LIME can be poor, because LIME does not take the graph structure information into account. Moreover, if the underlying model is highly nonlinear even in the locality of the prediction, such as that in the case of GNN models, a linear explanation model may not be able to produce faithful explanations.

In this paper, we propose another GNN explanation method based on LIME \cite{ribeiro2016should} in a nonlinear manner. More specifically, we propose GraphLIME, a model-agnostic and nonlinear approach for providing locally faithful explanations for GNN-based models in a subgraph, whose procedure with a toy explanation sample is shown briefly in Figure \ref{GraphLIME}. It samples $N$-hop network neighbors of the node being explained and locally captures the nonlinear dependency between features and predictions based on HSIC Lasso. GraphLIME is a simple yet effective interpretation method, which locally finds the most representative features as explanations in a nonlinear manner. Note that GraphLIME can be treated as a nonlinear graph-variant of the LIME method \cite{ribeiro2016should}, which considers perturbation near the node being explained and applies a linear interpretable model. Generally, Nonlinear explanation methods perform better than linear methods as shown in Figure 2. Through experiments on two real-world datasets, it is shown that the explanations of GraphLIME are of extraordinary degree and more descriptive in comparison to the existing explanation methods.

\section{Related Work}
In this section, we briefly review related work from two aspects. The first one is the recent development of GNNs, the second one is the review of interpretation methods for neural models. 
\subsection{Graph Neural Networks}
Recently, there have been many studies on extending deep learning approaches for graph data. GNNs \cite{scarselli2008graph} recursively incorporate the node feature information and structure information of a graph from the neighborhood into a neural network and obtain node embeddings for subsequent machine learning tasks. GraphSAGE \cite{hamilton2017inductive} is a spatial based and general inductive framework that can generate a node's embeddings by essentially assembling a node's neighborhood information from the node's local neighborhood, and its aggregation function must be invariant to permutations of node orderings. Novel control variate-based stochastic approximation algorithms for GCN \cite{chen2017stochastic} have been proposed, which utilize the historical activations of nodes as a control variate and allow sampling an arbitrarily small neighbor size. Graph Attention Network (GAT) \cite{velivckovic2017graph} is one type of GNNs, which seeks an aggregation function to fuse the neighboring nodes, random walks, and candidate models in graphs to learn a new representation. The main idea of GATs is to employ attention mechanisms that assign larger weights to the more important nodes, walks, or models. Gated Attention Network (GaAN) \cite{zhang2018gaan} is a new network architecture that uses a convolutional sub-network to compute a soft gate at each attention head to control its importance, and the gated attention can modulate the amount of attended content via the introduced gates. Graph Attention Model (GAM) \cite{lee2018graph} processes only a portion of the graph by adaptively selecting a sequence of "informative" nodes and is equipped with an external memory component, which allows it to integrate the information gathered from different parts of the graph. Dual Graph Convolutional Network \cite{zhuang2018dual} was proposed as a simple and scalable semi-supervised learning method for graph-structured data in which only a very small portion of the training data are labeled, and it jointly considers local and global consistency assumptions. 

%\makoto{DNN explanation (LIME, saliency based method, etc.) and GNN explanation}
\subsection{Interpretability and Explanations for Neural Networks}
There are two main families of models to provide interpretability and explanations for neural models. The first family of models focuses on finding a simple and proxy model to interpret the model being explained in a model-agnostic way. For example, LIME \cite{ribeiro2016should} has been proposed to explain the predictions of any classifier by learning an interpretable linear model locally around the prediction. Black Box Explanations through Transparent Approximations (BETA) \cite{lakkaraju2017interpretable} is another model-agnostic framework to explain the behavior of the model being explained by simultaneously optimizing for fidelity to the original model and interpretability of the explanation. DeepRED \cite{zilke2016deepred} was proposed to extract rules from deep neural networks and generate explanations for predictions, and ANN-DT \cite{schmitz1999ann} was proposed to extract binary decision trees from neural networks, it is an interpretable model to interpret the main model.

The second family of models focuses on the relevant aspects of computation in the neural model being explained. Erhan et al. \cite{erhan2009visualizing} proposed to inspect feature gradients to find good qualitative interpretations of high-level features represented by neural models. 
DeepLIFT (Deep Learning Important FeaTures) \cite{shrikumar2017learning} was proposed to enable interpretability of neural networks by comparing the activation of each neuron to its 'reference activation' and assigning contribution scores according to the difference. The method proposed in \cite{sundararajan2017axiomatic} uses two fundamental axioms and a standard gradient operator to attribute the prediction of a deep network to its input features. It is worth noting that these methods produce a "saliency map" \cite{zeiler2014visualizing}, which reveals the important features of input data. However, recent works have shown that saliency map will give rise to misleading results in some instances \cite{adebayo2018sanity} and can lead to issues such as gradient saturation in some cases. This issue is especially grievous for discrete inputs such as graph adjacency matrices. Therefore, these methods are not suitable for interpreting GNNs, and this is what we do in this paper. Instead of creating surrogate models, which are inherently interpretable, some methods has been proposed to identify the patterns of input data and find influential samples for relevant information \cite{koh2017understanding,yeh2018representer}. However, few studies have focused on the structural information of graphs with these kinds of interpretation methods, although the relations between nodes in the graph are critical to the predictions from deep models. 
Recently, Ying et al. \cite{ying2019gnnexplainer} proposed to utilize mutual information to find a subgraph with associated features for interpreting GNN models; however, this method depends on discrete optimization, and the optimization is hard and only consider the predicted label of the node being explained. The proposed GraphLIME method utilizes predicted labels from both the node being explained and its neighbors, which enables it to capture more local information around the node and present a finite number of features as explanations in an intuitional way.

\begin{figure}[t]
%\hspace*{-1.2em}
\centering
\centering
\includegraphics[width=12cm]{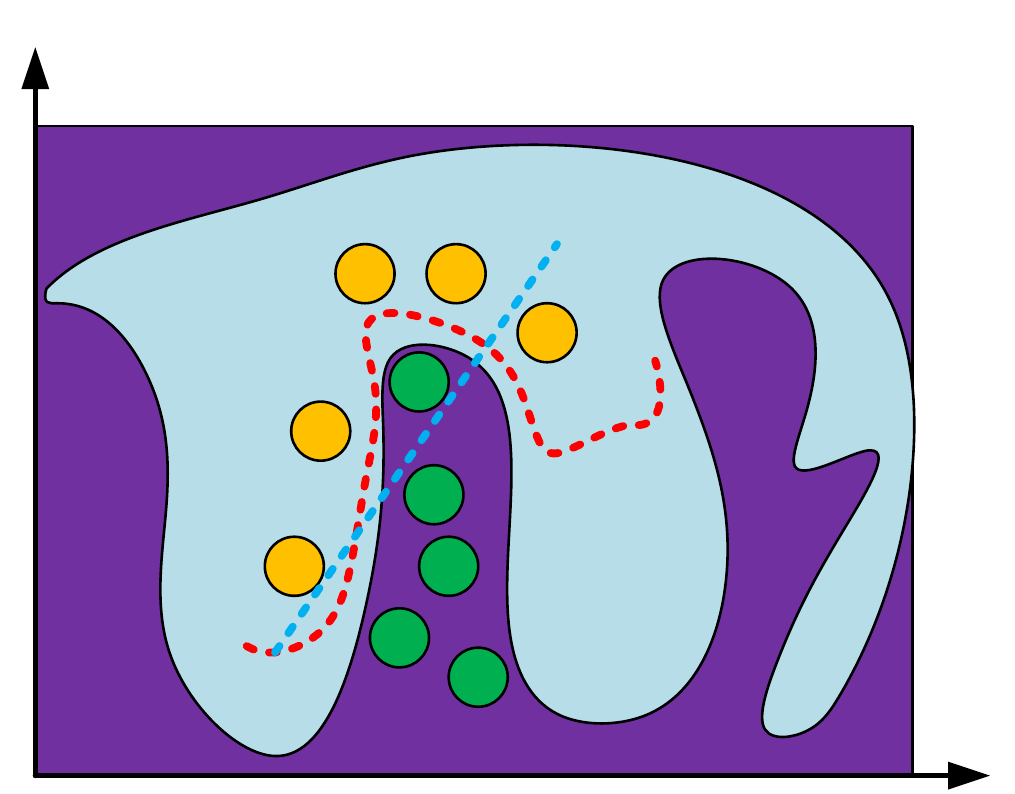}
%\caption{fig2}
\centering
\caption{A toy sample to present the advantage of nonlinear explanation model comparing to linear explanation model. Assuming that the purple and virescent background represent the GNN's complex decision function $f$, which is a black box to explanation model, the yellow and green filledCircle represent samples which belong to two different classes. When we use linear explanation model (the blue dashed line) to approximate the decision function $f$ locally, the linear model might make some mistakes due to its linearity, especially for the complex neural models. However, we can approximate the decision function $f$ precisely if we use a nonlinear explanation model (the red dashed line) and then identify some informative  features as explanations.}
\end{figure}

\section{Proposed Method: GraphLIME}
In this paper, we propose Local Interpretable Model Explanation for Graph Neural Networks (GraphLIME), which is a nonlinear model-agnostic explanation framework for GNNs. The goal of the proposed explanation framework is to identify a nonlinear interpretable model over the neighbors of the node being explained in a subgraph, which are locally faithful to the node's prediction, and interpret the GNN model, which is powerful but hard to explain in an intuitional manner. 
%\vspace{-.05in}
%\subsection{Understandable Data Representations and interpretable Models}
%Before we present our proposed explanation method, we need to know which pattern of manifestation for data is understandable to human. Understandable representations of data need to be made of features with practical significance, typically they are the original features extracted from real world. For example, in Boston House Price Dataset, each feature (e.g., Average Number of Rooms (RM), Capital Crime Rate by Town (CRIM)) represents the practical significance that can be understood by human. However, some data representations, such as word embedding, are incomprehensible for human.

%For some traditional machine learning algorithms such as logistic regression and decision tree that are interpretable, we can find most representative features by the values of coefficients or the branches and then interpret their predictions. However, when we fit the original data to the GNN model, the model will regroup the features and make the prediction in a nonlinear and complex manner which can not be understood intuitively by human, therefore, we can not identify which feature is more correlative with the prediction from GNN model. Our goal is to use the proposed explanation framework to build interpretable nonlinear model and then identify some original features as explanations that are locally faithful to the prediction.   
%\vspace{-.05in}

\subsection{Formulation of GraphLIME Explainer}
Formally, let $\mathcal{X} \subset \mathbb{R}^d$ be the domain of a vector $\boldx$. For each node $v_i$ in a graph, we have an associated feature vector $\boldx_i$, where each feature is extracted from the real world and can be understood by humans. We define an explanation as a model $g \in G$, where $G$ is a class of \emph{interpretable} models such as linear regressions, decision trees etc. In other words, a model $g \in G$ can provide intuitional interpretations in an interpretable manner. The explanation model class $G$ used in this paper is Hilbert-Schmidt Independence Criterion Lasso (HSIC Lasso) \cite{yamada2014high,yamada2018ultra,10.1093/bioinformatics/btz333}, which is a kernel based nonlinear interpretable feature selection algorithm. We will discuss more details about this in a later subsection. Let the domain of the explanation model $g \in G$ also be $\mathcal{X} \subset \mathbb{R}^d$, which means that $g$ acts over the original features that are understandable.

Let the GNN model being explained be denoted by $f:\mathbb{R}^d \rightarrow \mathbb{R}$. In a classification problem, $f(\boldx_i)$ denotes the probability (or binary indicator) that the instance $\boldx_i$ belongs to a certain class%\footnote{For multiple classes, we explain each class separately.}%, thus, $f(\boldx)$ is the prediction of the relevant class}
. Let $v$ denote the node whose prediction needs to be explained, $\boldX_n \in \mathbb{R}^{n \times d}$ represents the sampling information matrix that can capture the locality of the explained node $v$, where $n$ is the number of neighbors of $v$. Then, we obtain an interpretable model $g \in G$ using the local information matrix $\boldX_n$ of $v$ to approximate $f$, and generate the locally faithful explanations based on $g$. 

Given a GNN model $f$, the node $v$ being explained, the sampling local information matrix $\boldX_n$ of $v$, and an interpretable explanation model $g$, the explanations for the explained $v$ are obtained as follows:
\begin{equation}
   \zeta(v) \leftarrow  \argmin_{g \in G}  g(f,\boldX_n),
\end{equation}
where $\zeta(v)$ is the set of features as explanations of the node being explained $v$, it is generated based on the optimal explanation model $g$. 

Actually, different explanation models and sampling methods can be used in this formulation. Here, we focus on the nonlinear model for explanations, which can capture the nonlinear dependency between features and outputs and find the most representative features, and on the $N$-hop network neighbors sampling method to obtain the locally faithful explanations in a subgraph. Next, we will discuss the details about the sampling method and nonlinear explanation model of GraphLIME.

\subsection{Sampling for Local Exploration by $N$-hop Network Neighbors}
In a graph, a node's local information is decided by its neighbors in a subgraph; hence, the explainers for a GNN model in a graph should not only focus on the features of the given node but also the correlations between the features of its neighbouring nodes, because a given target node needs to aggregate information from its neighboring nodes and then make a prediction. In order to learn the local behavior of a GNN model being explained, we consider $N$-hop network neighbors to sample the neighboring nodes of a given node. A node $v$'s $N$-hop network is defined as the network formed by $v$ and the nodes whose distance from $v$ is within $N$ hops and links. By performing $N$-hop network sampling, we obtain:
$$\boldX_n=[\boldx_1, \boldx_2, ...\boldx_m]$$
where $\boldx_i$ is the corresponding associated feature vector of $v_i$, $v_i \in \calS_N $, $\calS_N$ is the set of $N$-hop neighbors of the explaining node $v$, and $m$ is the number of N-hop neighbors of $v_i$. Moreover, we can also obtain prediction $y_i=f(\boldx_i)$ for a given GNN model $f$, which can be used as the label in the explanation model $g$. We now present a nonlinear and interpretable algorithm as the explanation model in the proposed framework GraphLIME.

\subsection{Nonlinear Explanation Model: HSIC LASSO}
We consider a feature-wise kernelized nonlinear method called Hilbert-Schmidt Independence Criterion Lasso (HSIC Lasso) \cite{yamada2014high,yamada2018ultra,10.1093/bioinformatics/btz333} as the explanation model for GNN models. The HSIC Lasso is a "supervised" nonlinear feature selection method. Given a GNN model $f$, a node $v$ whose prediction needs to be explained, the neighbor set $\boldX_n$ of node $v$, and "supervised" paired $N$-hop neighboring nodes $\{(\boldx_i,{y}_i)\}_{i = 1}^n$, where $\boldx_i \in \boldX_n$, the probability $y_i=f(\boldx_i)$ that $x_i$ belongs to a certain class is used as the \emph{label} of the HSIC Lasso explanation model. For each prediction being explained, the HSIC Lasso optimization problem for an explanation model $g \in G$ is given as
\begin{equation}
    \label{eq:signedlasso}
  \begin{split}
  \min_{\boldbeta \in \mathbb{R}^d}&\frac{1}{2}\|\bar{\boldL}-\sum_{k=1}^{d} \beta_k\bar{\boldK}^{(k)}\|_F^2+\rho\| \boldbeta \|_1 \\
  &\text{s.t.} \quad \beta_1,\ldots,\beta_d\geq0,
  \end{split}
\end{equation}
where $\| \cdot \|_F$ is the Frobenius norm, $\rho \geq 0$ is the regularization parameter, $\|\cdot\|_1$ is the $\ell_1$ norm to enforce sparsity, $\bar{\boldL} = \boldH\boldL\boldH/\|\boldH\boldL\boldH\|_F$ is the normalized centered Gram matrix, $\boldL_{ij} = L({y}_i, {y}_j)$ is the kernel for the output,  $\boldH = \boldI_n -\frac{1}{n} \boldone_n\boldone_n^\top$ is the centering matrix, $\boldI_n$ is the $n$-dimension identity matrix, $\boldone_n$ is the $n$-dimension vector whose elements are all 1, $\bar{\boldK}^{(k)} = \boldH\boldK^{(k)}\boldH/\|\boldH\boldK^{(k)}\boldH\|_F$ is the normalized centered Gram matrix for the $k$-th feature, and $[\boldK^{(k)}]_{ij} = K(\boldx_i^{(k)}, \boldx_j^{(k)})$ is the kernel for the $k$-th dimensional input. In this paper, we use the Gaussian kernel for both input and the predictions of $\boldX_n$ from the given GNN model:
\begin{align*}
    K(\boldx_i^{(k)},\boldx_j^{(k)})&=\exp\left({-\frac{(\boldx_i^{(k)}-\boldx_j^{(k)})^2}{2\sigma_x^2}}\right),\\
L(y_i,y_j)&=\exp\left({-\frac{\|y_i-y_j\|_2^2}{2\sigma_y^2}}\right),
\end{align*}
where $\sigma_x$ and $\sigma_y$ are the Gaussian kernel widths. {Note that we consider the complete graph in this paper. However, we can explicitly take the local graph information into account by modifying $\boldK \circ \boldA \rightarrow \boldK$ and $ \boldL \circ \boldA \rightarrow \boldL$ respectively, where $\boldA \in \{0,1\}^{n \times n}$ is an adjacency matrix with self-loop and $\circ$ is the elementwise product.  }

We employ the nonnegative least angle regression \cite{efron2004least} to optimize the Eq.~\eqref{eq:signedlasso}, and then we can obtain the coefficient vector $\boldbeta$ and select the top-$K$ features as the explanations for the prediction of the node being explained based on it. 

\subsection{Interpretation of HSIC Lasso} %Provided
Here, we present the interpretation of the nonlinear explanation model HSIC Lasso. The HSIC Lasso contains the main concepts of minimum redundancy maximum relevancy (mRMR) \cite{peng2005feature}, which is a widely used classical supervised feature selection algorithm that can find non-redundant features with strong dependence on the output values. We can rewrite the first term of Eq.~\eqref{eq:signedlasso} as
\begin{align}
\label{eq:interpre}
  \begin{split}
     &\frac{1}{2}\|\bar{\boldL}-\sum_{k=1}^{d} \beta_k\bar{\boldK}^{(k)}\|_F^2\\
     &=\frac{1}{2}\!\!\sum_{k,m=1}^{d}\!\!\beta_k\beta_m \text{NHSIC}(\boldf_k,\boldf_m) 
     \!-\!\sum_{k=1}^{d}\!\!\beta_k \text{NHSIC}(\boldf_k,\boldy) + \frac{1}{2}\\
     %&\phantom{=}+\frac{1}{2}\text{NHSIC}(\boldy,\boldy),
  \end{split}
\end{align}
where $\boldf_k \in \mathbb{R}^n $ is the feature vector corresponding to the $k$-th feature,   $\text{NHSIC}(\boldf_k,\boldy)=\text{tr}(\bar{\boldK}^{(k)}\bar{\boldL})$ is the normalized variant of the empirical estimate of the Hibert-Schmidt independence criterion (HSIC) \cite{gretton2005measuring}, $\text{NHSIC}(\boldy,\boldy) = 1$ is constant,  and $\text{tr}(\cdot)$ is the trace operator. HSIC, which is based on a universal reproducing kernel such as the Gaussian kernel, is a non-negative function that estimates the independence between two random variables. A larger HSIC value indicates more dependency between the two variables, and it is zero if and only if the two random variables are statistically independent. The proof of Eq. (3) is left to Appendix.

In Eq.~\eqref{eq:interpre}, we ignore the value of $1/2$ because it is constant.  We consider the values of NHSIC$(\boldf_k,\boldy)$ and NHSIC$(\boldf_k,\boldf_m)$. For NHSIC$(\boldf_k,\boldy)$, if there is strong dependency between the $k$-th feature vector $\boldf_k$ and the output vector $\boldy$, the value of  NHSIC$(\boldf_k,\boldy)$ should be large and the corresponding  coefficient $\beta_k$ should also take a large value in order to minimize Eq.~\eqref{eq:signedlasso}. Meanwhile, if $\boldf_k$ is independent of $\boldy$, the value of NHSIC$(\boldf_k,\boldy)$ should be small so that $\beta_k$ tends to be eliminated by $l_1$-regularizer. This property can help select the most relevant features from the output vector $\boldy$.

For NHSIC$(\boldf_k,\boldf_m)$, if $\boldf_k$ and $\boldf_m$ are strongly dependent (i.e., redundant features), the value of NHSIC$(\boldf_k,\boldf_m)$ should be large and  either of the two coefficients $\beta_k$ and $\beta_m$ tends to be zero in order to minimize the Eq.~\eqref{eq:signedlasso}. This means that the redundant features will not be selected by the HSIC Lasso.

\begin{algorithm}[t]
	\renewcommand{\algorithmicrequire}{\textbf{Input:}}
	\renewcommand{\algorithmicensure}{\textbf{Output:}}
	\caption{Locally nonlinear Explanation: GraphLIME}
	\label{alg:1}
	\begin{algorithmic}[1]
		\REQUIRE GNN classifier $f$, Number of explanation features K
		\REQUIRE the graph $\mathcal{G}$, the node $v$ being explained
		\ENSURE $K$ explanation features
		\STATE $\boldX_n = N\_hop\_neighbor\_sample(v)$
		\STATE $\boldZ \leftarrow \{\}$
		\FORALL{$x_i \in \boldX_n$}
		\STATE $y_i=f(\boldx_i)$
		\STATE $\boldZ \leftarrow \boldZ \cup (\boldx_i, y_i) $
		\ENDFOR
		\STATE $\boldbeta \leftarrow \text{HSIC Lasso} (\boldZ)$ $\rhd$ with $x_i$ as features, $y_i$ as label
		\STATE $\zeta(v) \leftarrow$ Top-$K$ features as explanations based on $\boldbeta$
	\end{algorithmic}  
\end{algorithm}

In Algorithm 1, we summarize the model-agnostic and nonlinear explanation framework based on $N$-hop network sampling and HSIC Lasso for GNNs as a procedure, which we call GraphLIME.

\section{Experiments}
In this section, we present the experiments to investigate the performance of the proposed framework GraphLIME and the existing explanation methods for GNN models using two real-world datasets. In particular, we address the following three questions: (1) Does the explainer select the real informative features from the noisy graph data, (2) Do the explanations from the explainers guide the users in ascertaining trust in the predictions, and (3) Do the explanations help further in selecting a better classifier. All codes are implemented by Python and we use Intel(R) Xeon(R) CPU E5-2690 v4 @ 2.60GHz 264G Memory, and NVIDIA Corporation GP100GL [Tesla P100 PCIe].

\subsection{Setting}
We trained the GraphSAGE \cite{hamilton2017inductive} and GAT \cite{lee2018graph}, which are widely used GNN models, for the following explanation experiments. We performed simulated user experiments to evaluate the effectiveness of the proposed framework GraphLIME and other explanation methods. More specifically, we compared the proposed framework GraphLIME with the LIME \cite{ribeiro2016should} framework, which utilizes the perturbation method to sample data and train a linear explanation model lasso for selecting features as explanations according to the coefficients from the linear explanation model. In addition, we compared it with GNNexplainer~\cite{ying2019gnnexplainer}, which utilizes mutual information to find a subgraph with associated features for interpreting GNN models. We also compared it with a method based on the greedy procedure~\cite{martens2013explaining}, which greedily removes the most contributory features of the prediction until the prediction changes, and the random procedure, which randomly selects $K$ features as the explanations for the prediction being explained.

%In our experiments, we used two graph datasets, namely Cora and Pubmed. Cora is a scientific publication dataset, and each publication in the dataset is described by a 0/1-valued word vector indicating the absence/presence of the corresponding word from a dictionary consisting of 1433 unique words. Pubmed is a diabetes dataset, and each publication in the dataset is described by a TF/IDF weighted word vector from a dictionary consisting of 500 unique words. The statistical details of the Cora and Pubmed datasets are given in brief in Table \ref{tb:dataset}.  We sampled the 2-hop network neighbors of the node being explained and then randomly split the datasets into training sets (80\%) and testing sets (20\%).
In our experiments, we used two graph datasets, namely Cora and Pubmed. Cora and pubmed are two publication datasets, each feature indicates the absence/presence of the corresponding word in Cora and the TF/IDF value of the corresponding word in Pubmed. More statistical details of the Cora and Pubmed datasets are given in brief in Table \ref{tb:dataset}.  We sampled the 2-hop network neighbors of the node being explained and then randomly split the datasets into training sets (80\%) and testing sets (20\%).
 \begin{table}[t]
\caption{Statistics of Cora and Pubmed Datasets.\label{tb:dataset}}\smallskip
\centering
\normalsize
%\resizebox{.95\columnwidth}{!}{
\setlength{\tabcolsep}{7mm}{
\begin{tabular}{ccc}
\toprule
Datasets & Cora & Pubmed\\
\hline
\# of Classes & 7 & 3\\
\# of Nodes & 2708 &19717\\
%\hline
\# of Features & 1433 & 500\\
%\hline
\# of Links & 5429 & 44338\\
%\hline
\bottomrule
\end{tabular}
}
\label{table1}
\end{table}
 
 %In particular, we sample the 2-hop network neighbors of the being explained node, split the datasets into training sets (80\%) and testing sets (20\%), we perform all the experiments on two graph datasets (Cora and Pubmed).  

 %representative features as the explanation in GraphLIME. And we set the number of perturbation samples N=100 and use these perturbed samples to train linear explanation model (Lasso model in this paper) to select explanation features in original LIME model. For Greedy procedure we set the maximum removed features is $10$ and for Random we randomly select $10$ features as the explanation.
 \begin{figure}[t]
     \centering
     \subfigure[]{
\begin{minipage}[t]{0.5\linewidth}
\centering
\includegraphics[width=2.6in]{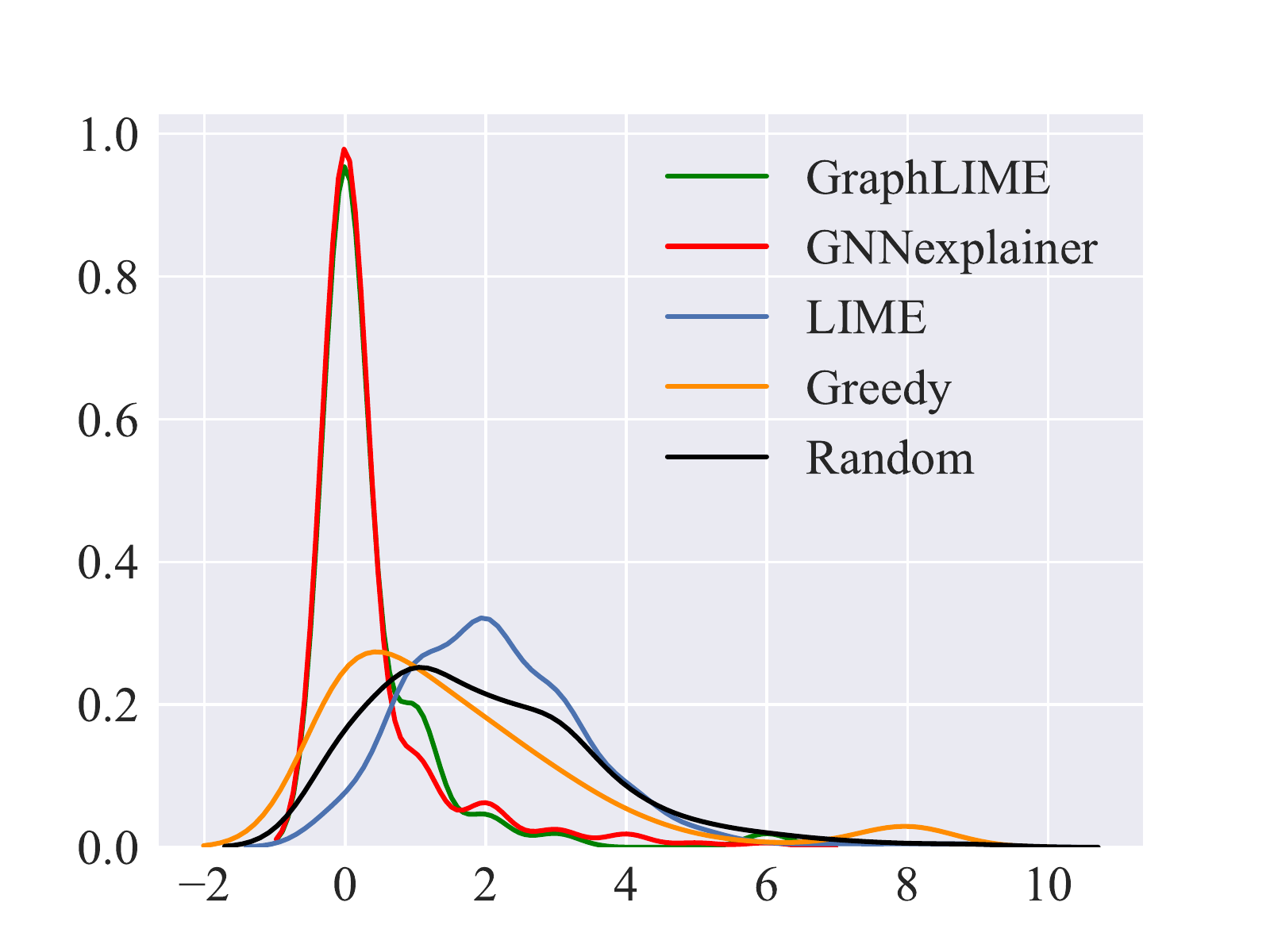}
%\caption{fig1}
\end{minipage}%
}%
\subfigure[]{
\begin{minipage}[t]{0.5\linewidth}
\centering
\includegraphics[width=2.6in]{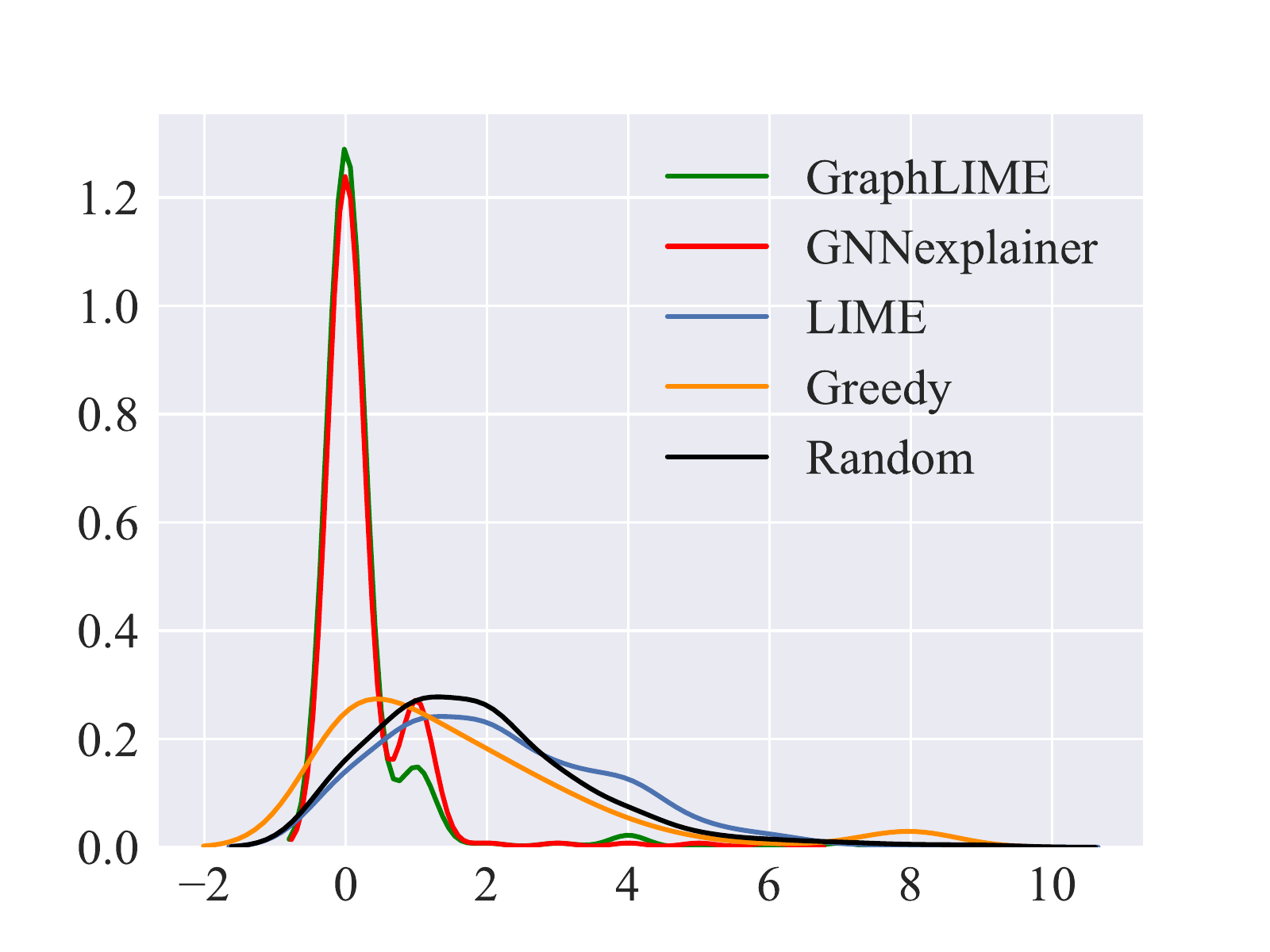}
%\caption{fig1}
\end{minipage}%
}%
\\
\subfigure[]{
\begin{minipage}[t]{0.5\linewidth}
\centering
\includegraphics[width=2.6in]{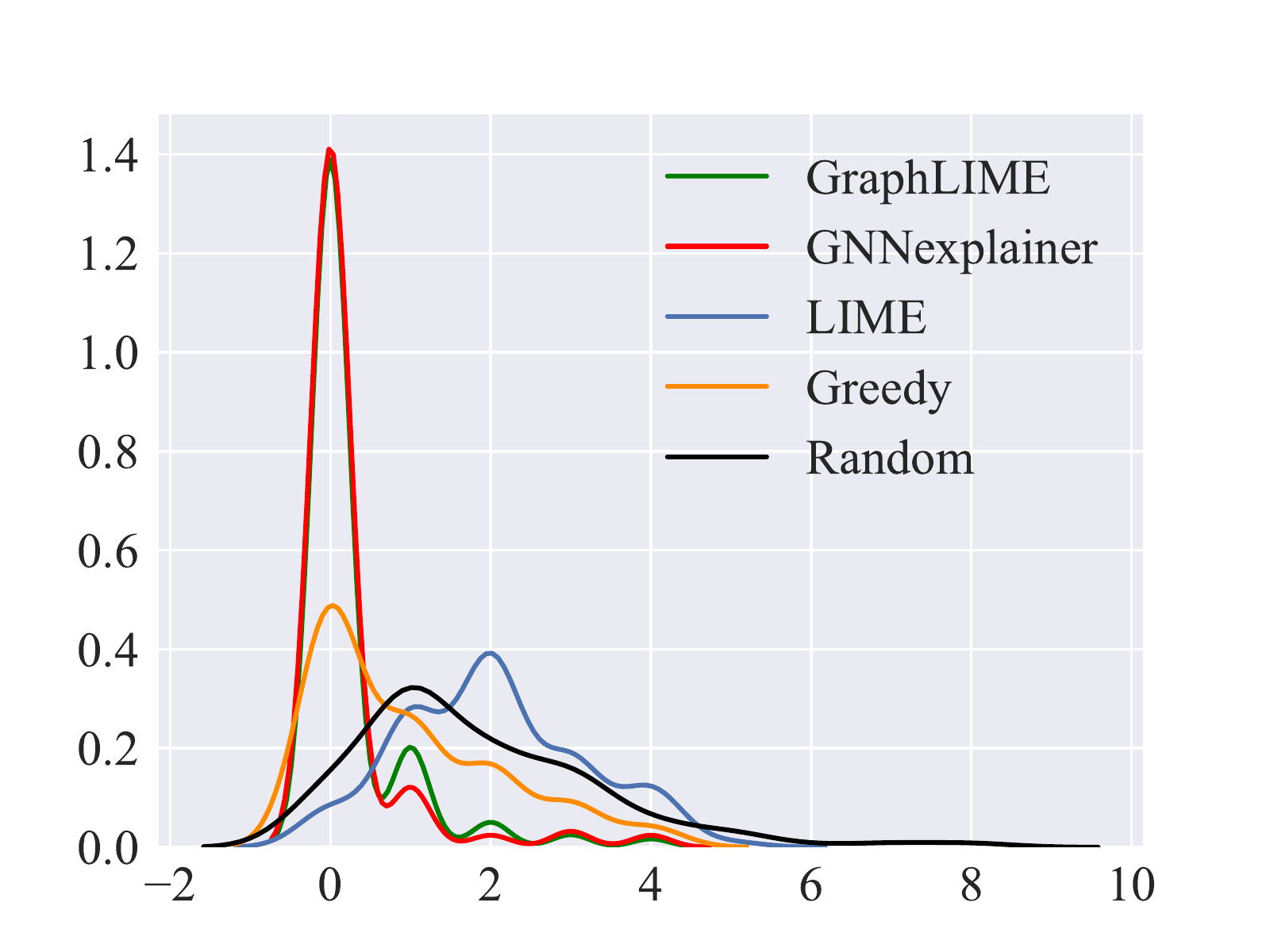}
%\caption{fig1}
\end{minipage}%
}%
\subfigure[]{
\begin{minipage}[t]{0.5\linewidth}
\centering
\includegraphics[width=2.6in]{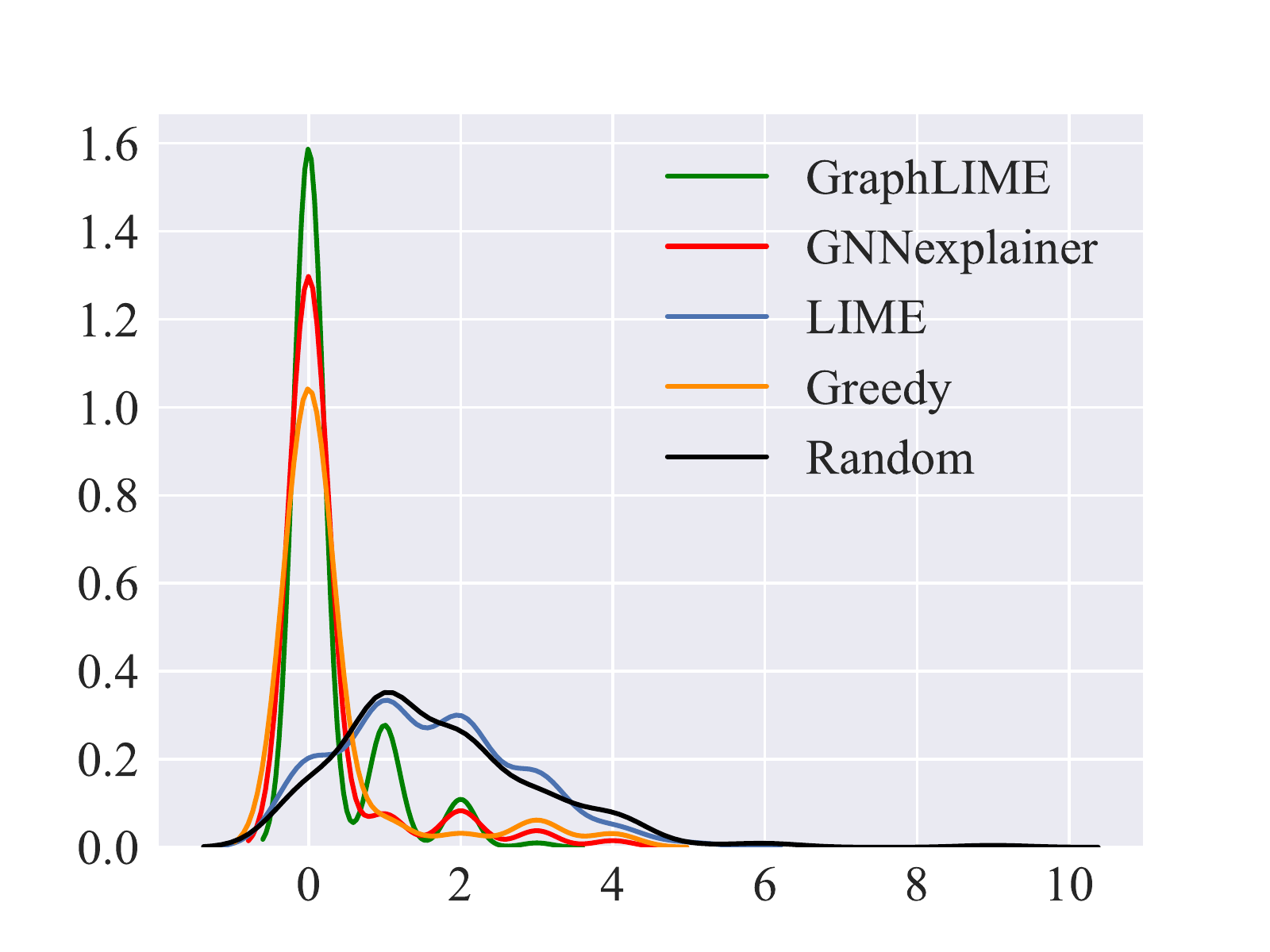}
%\caption{fig1}
\end{minipage}%
}%
\caption{Distribution of noisy features on Cora for GraphSAGE; (b) Distribution of noisy features on Pubmed for GraphSAGE; (c) Distribution of noisy features on Cora for GAT; (d) Distribution of noisy features on Pubmed for GAT.\label{fig:fig2}}
 \end{figure}

\begin{table*}[t]
\setlength\tabcolsep{2.5pt}
\begin{minipage}[t]{0.55\linewidth}
  \renewcommand{\arraystretch}{1.65}
  \centering
  \caption{Average F1-Score (\%) of trustworthiness for \\ different explainers on Cora for GraphSAGE. \label{tb:table2}}
  \label{table1}
  \centering
  {
  \begin{tabular}{ccccc}
  \toprule
         Method & $K$=10 & $K$=15 &$K$=20 &$K$=25  \\
         \hline
         Random & 40.4$\pm4.3$ &37.7$\pm4.6$ & 36.6$\pm4.3$ &36.0$\pm4.2$\\
         Greedy & 71.7$\pm2.9$ &71.6$\pm3.2$ & 71.5$\pm3.3$&71.5$\pm3.2$\\
         GNNexplainer &77.6$\pm3.1$ &71.5$\pm2.9$ &77.1$\pm1.4$ &76.5$\pm2.5$\\
         LIME &89.3$\pm1.8$& 89.7$\pm1.7$& 89.8$\pm1.1$&89.8$\pm1.1$\\
         GraphLIME& \textbf{95.3$\pm\textbf{0.5}$}& \textbf{95.6$\pm\textbf{0.7}$}& \textbf{95.4$\pm\textbf{0.7}$}&\textbf{95.4$\pm\textbf{0.6}$}\\
         \bottomrule
  \end{tabular}}
  \end{minipage}
%\\[10pt]
\begin{minipage}[t]{0.55\linewidth}
  \renewcommand{\arraystretch}{1.65}
  \caption{Average F1-Score (\%) of trustworthiness for different explainers on Pubmed for GraphSAGE.\label{tb:table3}}
  \label{table2}
  \centering
  {
  \begin{tabular}{ccccc}
   \toprule
         Method & $K$=10 & $K$=15 & $K$=20 & $K$=25  \\
         \hline
         Random & 21.9$\pm2.6$ &16.9$\pm2.1$ & 15.3$\pm1.7$ & 14.6$\pm1.6$\\
         Greedy & 63.5$\pm7.5$ &62.8$\pm7.3$ & 62.4$\pm7.3$&62.6$\pm7.2$\\
         GNNexplainer&79.5$\pm2.7$ & 77.6$\pm2.2$ &79.3$\pm3.7$&75.1$\pm1.6$\\
         LIME & 83.6$\pm0.8$& 84.0$\pm0.9$& 84.5$\pm0.9$&84.6$\pm0.9$\\
         GraphLIME& \textbf{92.5$\pm\textbf{0.8}$}& \textbf{92.3$\pm\textbf{0.8}$}& \textbf{91.6$\pm\textbf{0.9}$}&\textbf{91.5$\pm\textbf{0.8}$}\\
         \bottomrule
  \end{tabular}}
  \end{minipage}
  %\vspace{-0.1cm}%
  \\[15pt]
  \begin{minipage}[t]{0.55\linewidth}
  \renewcommand{\arraystretch}{1.65}
  \caption{Average F1-Score (\%) of trustworthiness for \\ different explainers on Cora for GAT.\label{tb:table3}}
  \label{table2}
  \centering
  {
  \begin{tabular}{ccccc}
   \toprule
         Method & $K$=10 & $K$=15 & $K$=20 & $K$=25  \\
         \hline
         Random & 37.1$\pm7.2$ &23.2$\pm5.9$ & 24.1$\pm6.8$ & 17.9$\pm5.3$\\
         Greedy & 72.9$\pm5.9$ &63.8$\pm5.3$ & 61.9$\pm5.8$&61.8$\pm5.7$\\
         GNNexplainer&73.7$\pm1.7$ & 79.0$\pm2.2$ &72.4$\pm1.8$&63.8$\pm2.3$\\
         LIME & 91.3$\pm1.8$& 87.6$\pm1.9$& 87.9$\pm2.1$&87.6$\pm2.4$\\
         GraphLIME& \textbf{96.1$\pm\textbf{1.1}$}& \textbf{96.2$\pm\textbf{0.9}$}& \textbf{95.1$\pm\textbf{1.4}$}&\textbf{95.3$\pm\textbf{1.4}$}\\
         \bottomrule
  \end{tabular}}
  \end{minipage}
%\\[10pt]
  \begin{minipage}[t]{0.55\linewidth}
  \renewcommand{\arraystretch}{1.65}
  \caption{Average F1-Score (\%) of trustworthiness for  different explainers on Pubmed for GAT.\label{tb:table3}}
  \label{table2}
  \centering
  {
  \begin{tabular}{ccccc}
   \toprule
         Method & $K$=10 & $K$=15 & $K$=20 & $K$=25  \\
         \hline
         Random & 36.2$\pm3.9$ &27.6$\pm1.3$ & 31.7$\pm5.0$ & 23.5$\pm5.8$\\
         Greedy & 55.8$\pm3.1$ &53.3$\pm3.6$ & 51.4$\pm2.7$&52.4$\pm2.6$\\
         GNNexplainer&70.8$\pm2.9$ &69.4$\pm2.1$ &65.6$\pm2.5$&61.7$\pm1.2$\\
         LIME & 79.1$\pm2.6$& 76.4$\pm2.9$& 70.3$\pm3.0$&71.3$\pm3.0$\\
         GraphLIME& \textbf{92.1$\pm\textbf{1.1}$}& \textbf{91.2$\pm\textbf{1.1}$}& \textbf{92.0$\pm\textbf{1.2}$}&\textbf{92.0$\pm\textbf{1.3}$}\\
         \bottomrule
  \end{tabular}}
  \end{minipage}
\end{table*}

\subsection{Does the explanation framework filter useless features?}
In the first simulated user experiment, we investigated whether GraphLIME could filter useless features and select informative features as the explanations. 
For this, we compared the frequency of samples on different number of selected "noisy" features over different explanation frameworks to compare their abilities of denoising data. 

%\vspace{-.06in}
Concretely, we artificially and randomly added 10 "noisy" features in each sample's feature vector and then trained a GraphSAGE model or GAT model whose test accuracy was more than 80\%. Thus we obtained the useless set of features (the 10 "noisy features") for the trained model. Finally, we produced explanations on 200 test samples for each explanation framework and compared their performance in terms of the frequency distribution of samples on different number of noisy features. 
%The goal of the experiment is to evaluate which explanation method can filter useless features and select real informative features as explanations. So . We test 200 instances for each dataset, select 10 features as the explanations and run 100 rounds to compute the the averaged frequency over different number of noisy features.

We used kernel density estimation (KDE) to plot the frequency distributions of samples on different number of noisy features for five different explanation frameworks on Cora and Pubmed and used Gaussian kernel to fit the shape of distributions. The distribution is shown in Figure \ref{fig:fig2}. It can be seen that the number of noisy features selected by the proposed framework GraphLIME and GNNexplainer are in general less than that selected by the other explanation frameworks. For GraphLIME and GNNexplainer, the frequency of samples on different number of selected noisy features was around 0. This means that the GraphLIME and GNNexplainer frameworks rarely select useless features as explanations, which is very useful when the graph data has a large amount of noise. LIME is not capable of ignoring the noisy features, and the distribution of LIME is mainly around 1 to 4. For Random procedure, the frequency distribution is similar to that of LIME. The Greedy explanation procedure is just slightly better than LIME and Random, but not comparable with GraphLIME and GNNexplainer. The results demonstrate that the proposed framework can denoise graph data for the GNN model when the data contains a large amount of noise and that it is more capable of finding informative features as explanations.

\subsection{Do I trust this prediction?}
The prediction made by a classifier model might not be credible; therefore, it is important for an explanation framework that the explanations can aid users in deciding whether a prediction is trustworthy. In this experiment, we compared this kind of ability for different explanation frameworks. 

Firstly, we randomly selected 30\% of the features as "untrustworthy" features. Then, we trained a GraphSAGE or GAT classifier and obtained the predictions on testing samples. We assumed that the users could identify these "untrustworthy" features and that they would not want these features as explanations. Secondly, we developed \emph{oracle} "trustworthiness" by labeling predictions from the classifier as "untrustworthy" if the prediction changed after removing those "untrustworthy" features from the samples (because that meant the prediction was mainly decided by those "untrustworthy" features), and "trustworthy" otherwise. We regard \emph{oracle} "trustworthiness" as the true label to decide whether a prediction is credible. Thirdly, for GraphLIME and LIME, we assumed that the simulated users regard predictions from the classifier as "untrustworthy" if the prediction made by another approximation linear model changes when all "untrustworthy" features that appear in the explanations are removed. For GNNexplainer, Greedy, and Random explainer, the prediction was deemed untrustworthy by the users if any "untrustworthy" features appeared in their explanations. Finally, we compared the decisions made by the simulated users with the \emph{oracle} "trustworthiness".

\begin{figure}[t]
%\hspace*{-1.2em}
\centering
\subfigure[Cora  for GraphSAGE]{
\begin{minipage}[t]{0.5\linewidth}
\centering
\includegraphics[width=2.6in]{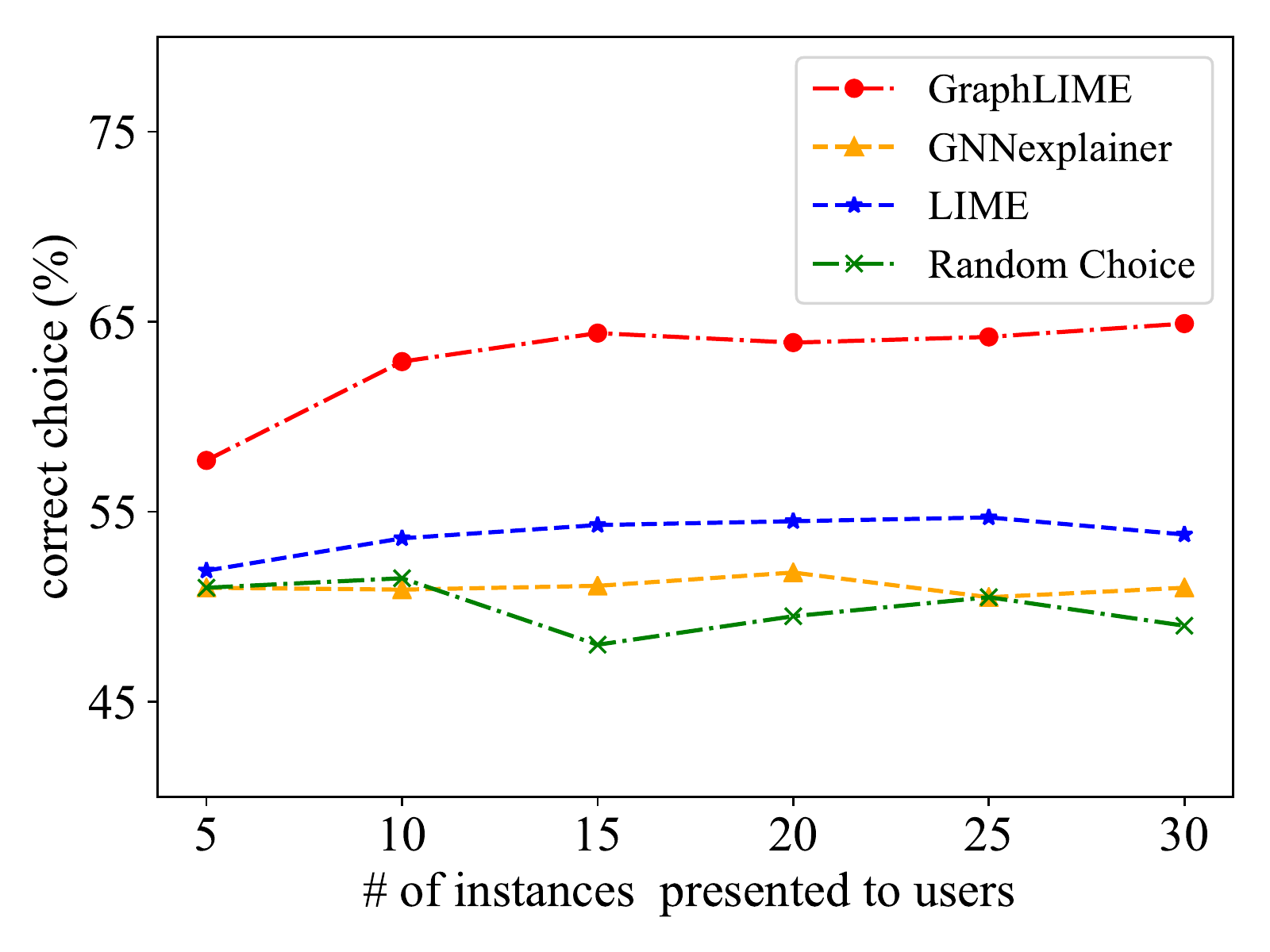}
\end{minipage}%
}%
\subfigure[Pubmed  for GraphSAGE]{
\begin{minipage}[t]{0.5\linewidth}
\centering
\includegraphics[width=2.6in]{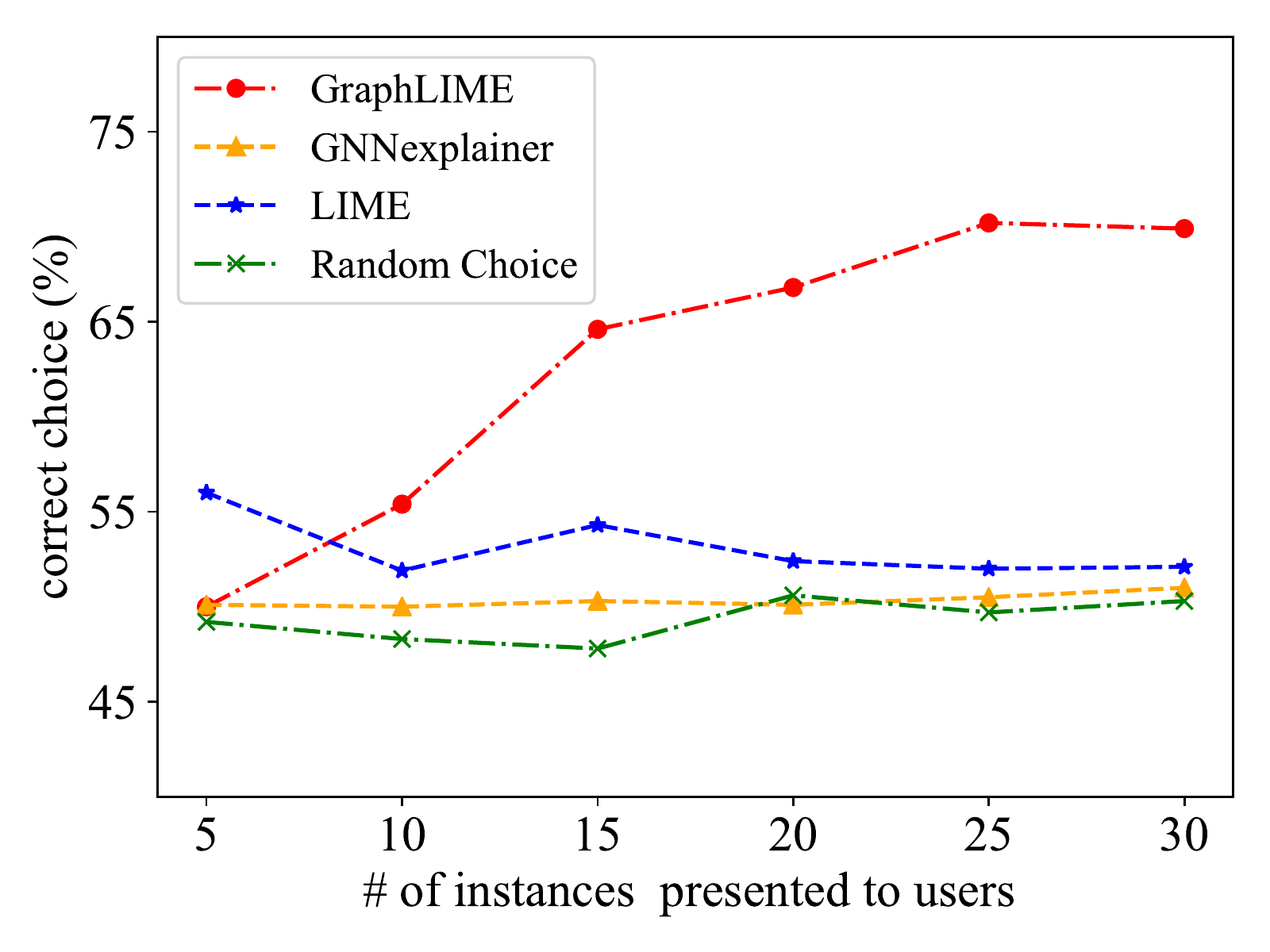}
%\caption{fig2}
\end{minipage}%
}%

\subfigure[Cora  for GAT]{
\begin{minipage}[t]{0.5\linewidth}
\centering
\includegraphics[width=2.6in]{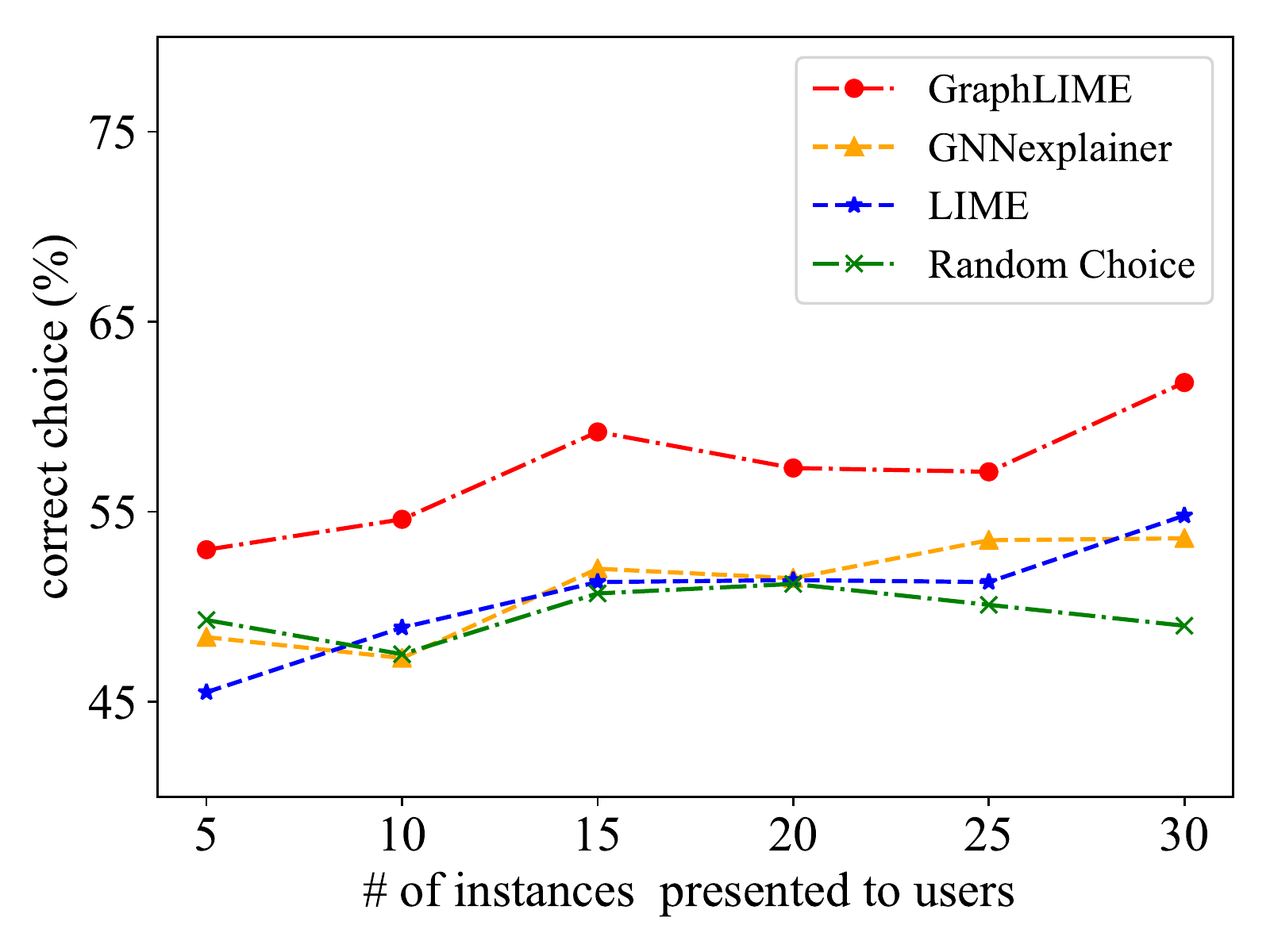}
%\caption{fig1}
\end{minipage}%
}%
\subfigure[Pubmed  for GAT]{
\begin{minipage}[t]{0.5\linewidth}
\centering
\includegraphics[width=2.6in]{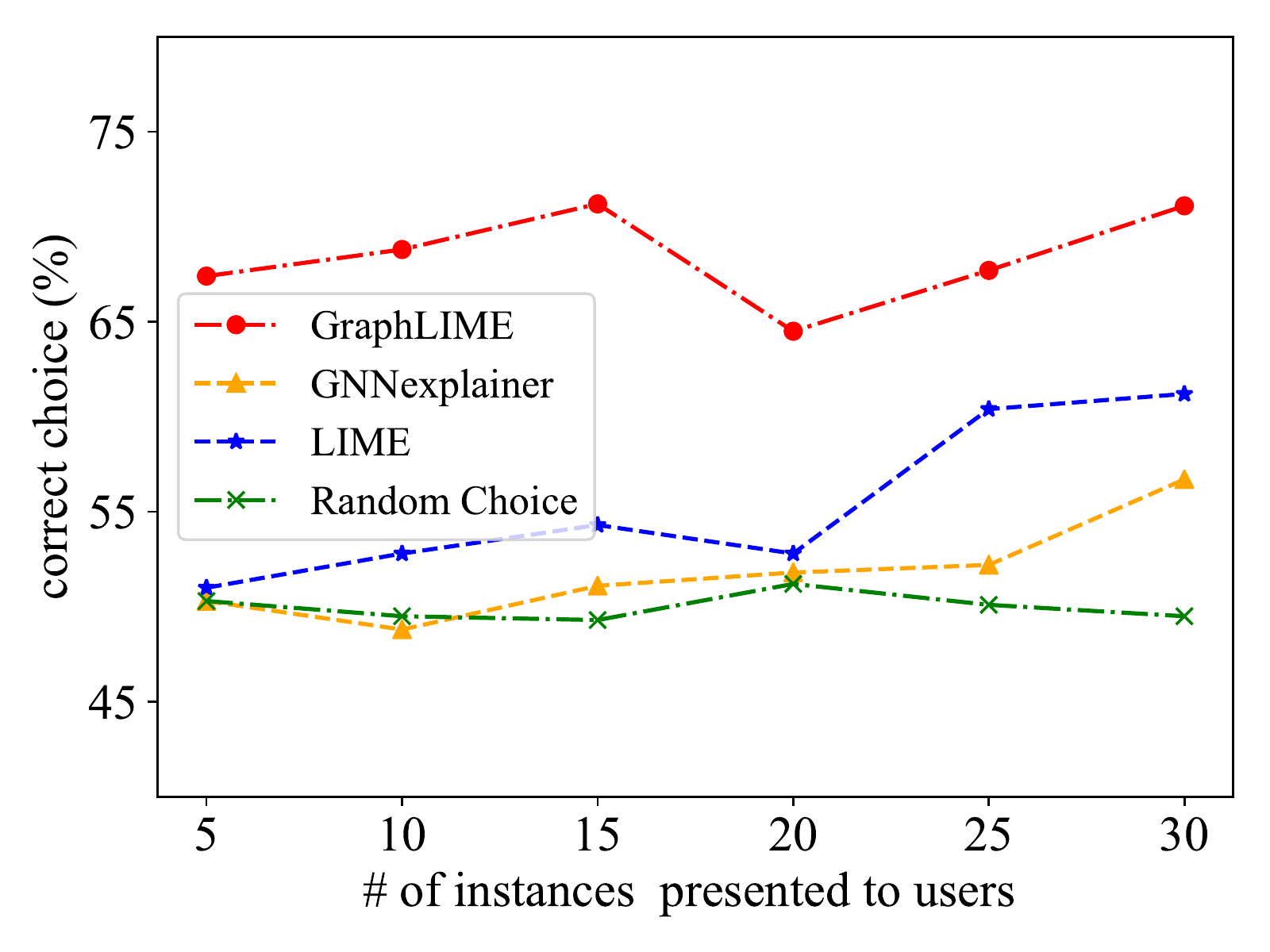}
%\caption{fig2}
\end{minipage}%
}%%
\centering
\caption{Choose the better one among the two classifiers, as the number of instances presented to users $B$ is varied for four explanation methods. Averages over 200 rounds.\label{acc_figure}}
\vspace{-.1in}
\end{figure}

We set the number of selected features as $K=10, 15, 20, 25$ and reported the averaged F1-Score on the trustworthy predictions for each explanation framework over 100 rounds. The results in Table \ref{tb:table2} and \ref{tb:table3} show that GraphLIME is superior to the other explanation methods on the two datasets. The lower F1-Score of the other explanation methods indicate that they achieved a lower precision (i.e., trusting too many predictions) or a lower recall (i.e., mistrusting predictions more than they should), while the higher F1-Score of the proposed framework GraphLIME indicates that it achieved both high precision and high recall.

\subsection{Does the explainer help to identify the better model?}
In the final simulated user experiment, we explored whether the explanation framework could be used for model selection and for guiding users to choose the \emph{better} one among two different GNN classifiers. We compared the performance of different frameworks in selecting the \emph{better} classifier by reporting the accuracy of selecting the real better classifier.

In order to simulate a situation where the models use not only informative features in the real world but also ones that introduce spurious relations, We artificially added 10 noisy features to the data, and marked them as untrustworthy features. We created pairs of classifiers by repeatedly training pairs of GraphSAGE or GAT classifiers until their training accuracy and test accuracy were both above 70\% and the difference in their test accuracy was more than 5\%. Then, we obtained explanations of the samples being explained by the explanation framework on two competing classifiers and recorded the number of untrustworthy features appearing in the explanations for the two classifiers. Note that the \emph{better} classifier should have fewer untrustworthy features in its explanations; therefore, we selected the classifier with fewer untrustworthy features as the better classifier and compared this choice with the real better classifier with higher test accuracy.

%Then we use a explainer to obtain the explanations for each instance's predictions from two classifiers. Finally we need to identify \emph{better} classifier (the one with higher test set accuracy) based on the explanations from the used explanation method.

%he goal of this experiment is to verify whether the proposed framework can guide users to identify better models based on explanations from explainers. Specifically, We obtain explanations of each instance on two competing classifiers and record the number of untrustworthy features appearing in the explanations for the two classifiers. Note that the \emph{better} classifier should have fewer untrustworthy features, therefore, we select the classifier with fewer untrustworthy features as the better classifier, and compare this choice to the real better classifier with higher test set accuracy. 

The goal of this experiment was to verify whether the explanation framework could guide users to identify better models based on explanations from the explanation framework. Considering that users may not have time to examine a large number of samples, let $B$ donate the number of samples being explained, which they are willing to look at in order to identify the better classifier. We use Submodular Pick \cite{ribeiro2016should} to select $B$ samples and present these samples to the users for examining, Submodular Pick will be detailed in the Appendix. 

We present the accuracy of selecting the real better classifier as $B$ varies from 5 to 30, averaged over 200 rounds, in Figure \ref{acc_figure}. We compared the performance of GraphLIME, LIME, GNNexplainer, and omitted Greedy and Random because they could not produce useful explanations as shown in the previous two experiments. Moreover, we also plotted the accuracy curve for random choice. These results demonstrate that the proposed method GraphLIME can be used to perform model selection and that it outperforms LIME and GNNexplainer, which are only slightly better than the random choice. And it is worthy to note that the performance of GraphLIME also improved With the increase in number of presented instances.

\section{Conclusion}
Most of the explanation methods for neural models are designed for general neural networks, while only few works exist for GNNs. In this paper, we presented a model-agnostic local interpretable explanation framework for GNN, which we call GraphLIME. It is able to leverage the feature information of the $N$-hop network neighbors of the node being explained and their predicted labels in a local subgraph, utilize the Hilbert-Schmidt Independence Criterion Lasso (HSIC Lasso), which is a nonlinear interpretable model for capturing the nonlinear dependency between features and predicted outputs, and produce finite features as the explanations for a particular prediction. Experiments on two real-world graph datasets for two kinds of GNN models demonstrated the effectiveness of the proposed framework. It could filter noisy features and select the real informative features, guide users in ascertaining trust in predictions, and help users to identify the better classifier. 

In the future, we would mainly focus on two aspects of the explanation for GNN models. Firstly, we will investigate how to extend the proposed GraphLIME so that it is able to explain the structural graph patterns and find the most important subgraph structure that influences the predictions from the GNN model. Secondly, the current framework can only explain the prediction for a single target node; therefore, we will investigate how to provide an explanation for a set of nodes, where all the nodes belong to one class from a given GNN model in a nonlinear manner.
\\[300pt]
\section*{Appendix}
\subsection*{A. Proof of Eq. (3)}

According to the definition and proof of HSIC \cite{gretton2005measuring} \cite{yamada2018ultra}, the definition of normalized  variant  of  the  empirical  estimate NHSIC is as follows:

\begin{align}
\label{eq:interpre}
  \begin{split}
    &\text{NHSIC}(\boldf_k,\boldy)=\text{tr}(\bar{\boldK}^{(k)}\bar{\boldL})\\
    &\text{NHSIC}(\boldf_k,\boldf_m)=\text{tr}(\bar{\boldK}^{(k)}\bar{\boldK}^{(m)})
  \end{split}
\end{align}
According to $\|\boldA\|_F^2=\text{tr}(AA^T)=\text{tr(AA)}$,where $A$ is  symmetric matrix, we can obtain the derivation as following:
\begin{small}
\begin{align}
  \begin{split}
     &\frac{1}{2}\|\bar{\boldL}-\sum_{k=1}^{d} \beta_k\bar{\boldK}^{(k)}\|_F^2\\
    &=\frac{1}{2}\text{tr}((\bar{\boldL}-\sum_{k=1}^{d}\beta_k \bar{K}^{(k)})\cdot(\bar{L}^T-\sum_{k=1}^{d}\beta_k \bar{K}^{(k)^T}))\\
     &=\frac{1}{2}\text{tr}((\bar{\boldL}-\sum_{k=1}^{d}\beta_k \bar{K}^{(k)})\cdot(\bar{L}^T-\sum_{k=1}^{d}\beta_k \bar{K}^{(k)}))\\
     &=\frac{1}{2}\text{tr}(\bar{L}\bar{L}^T-2\sum_{k=1}^{d}\beta_k\bar{L}\bar{K}^{(k)}+\sum_{k,m=1}^{d}\beta_k\beta_m\bar{K}^{(k)}\bar{K}^{(m)})\\
     &=\frac{1}{2}\text{tr}(\sum_{k,m=1}^{d}\beta_k\beta_m\bar{K}^{(k)}\bar{K}^{(m)})-\text{tr}(\sum_{k=1}^{d}\beta_k\bar{L}\bar{K}^{(k)})+\frac{1}{2}\text{tr}(\bar{L}\bar{L}^T)\\
     &=\frac{1}{2}\sum_{k,m=1}^{d}\beta_k\beta_m\text{tr}(\bar{K}^{(k)}\bar{K}^{(m)})-\sum_{k=1}^{d}\beta_k\text{tr}(\bar{L}\bar{K}^{(k)})+\frac{1}{2}\text{tr}(\bar{L}\bar{L}^T)\\
     &=\frac{1}{2}\!\!\sum_{k,m=1}^{d}\!\!\beta_k\beta_m \text{NHSIC}(\boldf_k,\boldf_m)
     \!\!-\!\!\sum_{k=1}^{d}\!\!\beta_k \text{NHSIC}(\boldf_k,\boldy) \!\!+\!\! \frac{1}{2}\text{NHSIC}(\boldy,\boldy)\\
     &=\frac{1}{2}\!\!\sum_{k,m=1}^{d}\!\!\beta_k\beta_m \text{NHSIC}(\boldf_k,\boldf_m)
     \!\!-\!\!\sum_{k=1}^{d}\!\!\beta_k \text{NHSIC}(\boldf_k,\boldy)\!\! +\!\! \frac{1}{2}\\
     %&\phantom{=}+\frac{1}{2}\text{NHSIC}(\boldy,\boldy),
  \end{split}
\end{align}
\end{small}
\subsection*{B. Submodular Pick}
Although the explanations from multiple instances can provide the powerful insight for users, users may not be willing to spend much time examining a great number of explanations. It's necessary to select a subset of instances advisably for the users to inspect. Here we consider a pick-step method called Submodular Pick\cite{ribeiro2016should}, which considers the global importance and diversity of each explanation component (the selected feature as explanation).

Given an instance set $X=\{\boldx_1,\boldx_2,\cdot\cdot\cdot,\boldx_n\}$, we can obtain the corresponding explanations $\zeta(v_i)$ and explanation model $g_i$ with coefficient vector $\boldbeta_i$ for certain instance $\boldx_i$ by GraphLIME. We construct explanation matrix $\boldW \in \mathbb{R}^{n \times d}$ that represents the local importance of each explanation component for instances, where
\[\boldW_{ij}=\begin{cases}

\boldbeta_i(j)&\text{, if $j \in \zeta(v_i)$}\\

0&

\text{otherwise}.

\end{cases}\]
$\boldbeta_i(j)$ is the corresponding coefficient value for feature $j$. Let the $I_j$ donates the global importance for component $j$, it is computed as follows:
$$
        I_j = \sqrt{\sum_{i=1}^n\boldW_{ij}}
$$
Intuitively, we want feature $j$ have higher global importance $I_j$ if it appears in many explanations of different instances.

Meanwhile, we want to pick instances which are not redundant, that means we want to avoid selecting similar instances to users. We need to select a set of instances such that their explanation components can cover all the features to the greatest extent to maintain the diversity of selected instances. We define the coverage score function $s$ as shown in Eq. (6) given explanation matrix $\boldW$, global importance $\boldI$ and the instance set $V$:
\begin{align}
    \begin{split}
        s(V,\boldW,\boldI) = \sum_{j=1}^d\mathbbm{1}_{[\exists i \in  V:\boldW_{ij}>0]}I_j
    \end{split}
\end{align}
Where $\mathbbm{1}$ is indicator function. With the definition above, the objective function of Submodular Pick can be written as follows:
\begin{align}
    \begin{split}
        Pick(\boldW,\boldI) =  \mathop{\arg\max}_{V,|V| = B} s(V,\boldW,\boldI)
    \end{split}
\end{align}
Where $B$ is the number of the instances users are willing to inspect. The problem of Eq. (7) is NP-hard, so we utilize a greedy algorithm that adds instance iteratively to get the instance subset $V$. The procedure of Submodular Pick is outlined in Algorithm 2.
\begin{algorithm}[t]
	\renewcommand{\algorithmicrequire}{\textbf{Input:}}
	\renewcommand{\algorithmicensure}{\textbf{Output:}}
	\caption{Submodular Pick}
	\label{alg:2}
	\begin{algorithmic}[1]
		\REQUIRE Instance set $X$, explanation matrix $W$, $B$
		\ENSURE Selected instance subset $V$ 
		\FORALL{$x_i \in \boldX$}
		\STATE $W_i \leftarrow GraphLIME(x_i)$  $\rhd$ Using Algorithm 1
		\ENDFOR
		\FORALL{$j \in \{1,...,d\}$}
		\STATE $I_j = \sqrt{\sum_{i=1}^n\boldW_{ij}}$
		\ENDFOR
		\STATE $V = \{ \}$
		\WHILE{$|V| < B$}
		\STATE $V = V \cup \mathop{\arg\max}_i c(V \cup \{i\}, \boldW, \boldI)$
		\ENDWHILE
		\STATE return $V$
	\end{algorithmic}  
\end{algorithm}
\bibliographystyle{unsrt}
\bibliography{ijcai20}

\begin{thebibliography}{10}

\bibitem{liu2017survey}
Weibo Liu, Zidong Wang, Xiaohui Liu, Nianyin Zeng, Yurong Liu, and Fuad~E
  Alsaadi.
\newblock A survey of deep neural network architectures and their applications.
\newblock {\em Neurocomputing}, 234:11--26, 2017.

\bibitem{lakkaraju2017interpretable}
Himabindu Lakkaraju, Ece Kamar, Rich Caruana, and Jure Leskovec.
\newblock Interpretable \& explorable approximations of black box models.
\newblock {\em arXiv preprint arXiv:1707.01154}, 2017.

\bibitem{ribeiro2016should}
Marco~Tulio Ribeiro, Sameer Singh, and Carlos Guestrin.
\newblock Why should i trust you?: Explaining the predictions of any
  classifier.
\newblock In {\em Proceedings of the 22nd ACM SIGKDD international conference
  on knowledge discovery and data mining}, pages 1135--1144. ACM, 2016.

\bibitem{chen2018learning}
Jianbo Chen, Le~Song, Martin~J Wainwright, and Michael~I Jordan.
\newblock Learning to explain: An information-theoretic perspective on model
  interpretation.
\newblock {\em In Proceedings of the 35th International Conference on Machine
  Learning}, 2018.

\bibitem{lundberg2017unified}
Scott~M Lundberg and Su-In Lee.
\newblock A unified approach to interpreting model predictions.
\newblock In {\em Advances in neural information processing systems}, pages
  4765--4774, 2017.

\bibitem{sundararajan2017axiomatic}
Mukund Sundararajan, Ankur Taly, and Qiqi Yan.
\newblock Axiomatic attribution for deep networks.
\newblock In {\em International Conference on Machine Learning}, pages
  3319--3328. JMLR. org, 2017.

\bibitem{koh2017understanding}
Pang~Wei Koh and Percy Liang.
\newblock Understanding black-box predictions via influence functions.
\newblock In {\em International Conference on Machine Learning}, pages
  1885--1894. JMLR. org, 2017.

\bibitem{yeh2018representer}
Chih-Kuan Yeh, Joon Kim, Ian En-Hsu Yen, and Pradeep~K Ravikumar.
\newblock Representer point selection for explaining deep neural networks.
\newblock In {\em Advances in neural information processing systems}, pages
  9291--9301, 2018.

\bibitem{ying2019gnnexplainer}
Zhitao Ying, Dylan Bourgeois, Jiaxuan You, Marinka Zitnik, and Jure Leskovec.
\newblock Gnnexplainer: Generating explanations for graph neural networks.
\newblock In {\em Advances in neural information processing systems}, pages
  9240--9251, 2019.

\bibitem{scarselli2008graph}
Franco Scarselli, Marco Gori, Ah~Chung Tsoi, Markus Hagenbuchner, and Gabriele
  Monfardini.
\newblock The graph neural network model.
\newblock {\em IEEE Transactions on Neural Networks}, 20(1):61--80, 2008.

\bibitem{hamilton2017inductive}
Will Hamilton, Zhitao Ying, and Jure Leskovec.
\newblock Inductive representation learning on large graphs.
\newblock In {\em Advances in neural information processing systems}, pages
  1024--1034, 2017.

\bibitem{chen2017stochastic}
Jianfei Chen, Jun Zhu, and Le~Song.
\newblock Stochastic training of graph convolutional networks with variance
  reduction.
\newblock {\em In Proceedings of the 35th International Conference on Machine
  Learning}, 2017.

\bibitem{velivckovic2017graph}
Petar Veli{\v{c}}kovi{\'{c}}, Guillem Cucurull, Arantxa Casanova, Adriana
  Romero, Pietro Li{\`{o}}, and Yoshua Bengio.
\newblock {Graph Attention Networks}.
\newblock {\em In International Conference on Learning Representations}, 2018.

\bibitem{zhang2018gaan}
Jiani Zhang, Xingjian Shi, Junyuan Xie, Hao Ma, Irwin King, and Dit-Yan Yeung.
\newblock Gaan: Gated attention networks for learning on large and
  spatiotemporal graphs.
\newblock {\em arXiv preprint arXiv:1803.07294}, 2018.

\bibitem{lee2018graph}
John~Boaz Lee, Ryan Rossi, and Xiangnan Kong.
\newblock Graph classification using structural attention.
\newblock In {\em Proceedings of the 22nd ACM SIGKDD international conference
  on knowledge discovery and data mining}, pages 1666--1674. ACM, 2018.

\bibitem{zhuang2018dual}
Chenyi Zhuang and Qiang Ma.
\newblock Dual graph convolutional networks for graph-based semi-supervised
  classification.
\newblock In {\em WWW}, pages 499--508. International World Wide Web
  Conferences Steering Committee, 2018.

\bibitem{zilke2016deepred}
Jan~Ruben Zilke, Eneldo~Loza Menc{\'\i}a, and Frederik Janssen.
\newblock Deepred--rule extraction from deep neural networks.
\newblock In {\em International Conference on Discovery Science}, pages
  457--473. Springer, 2016.

\bibitem{schmitz1999ann}
Gregor~PJ Schmitz, Chris Aldrich, and Francois~S Gouws.
\newblock Ann-dt: an algorithm for extraction of decision trees from artificial
  neural networks.
\newblock {\em IEEE Transactions on Neural Networks}, 10(6):1392--1401, 1999.

\bibitem{erhan2009visualizing}
Dumitru Erhan, Yoshua Bengio, Aaron Courville, and Pascal Vincent.
\newblock Visualizing higher-layer features of a deep network.
\newblock {\em University of Montreal}, 1341(3):1, 2009.

\bibitem{shrikumar2017learning}
Avanti Shrikumar, Peyton Greenside, and Anshul Kundaje.
\newblock Learning important features through propagating activation
  differences.
\newblock In {\em International Conference on Machine Learning}, pages
  3145--3153. JMLR. org, 2017.

\bibitem{zeiler2014visualizing}
Matthew~D Zeiler and Rob Fergus.
\newblock Visualizing and understanding convolutional networks.
\newblock In {\em European Conference on Computer Vision}, pages 818--833.
  Springer, 2014.

\bibitem{adebayo2018sanity}
Julius Adebayo, Justin Gilmer, Michael Muelly, Ian Goodfellow, Moritz Hardt,
  and Been Kim.
\newblock Sanity checks for saliency maps.
\newblock In {\em Advances in Neural Information Processing Systems}, pages
  9505--9515, 2018.

\bibitem{yamada2014high}
Makoto Yamada, Wittawat Jitkrittum, Leonid Sigal, Eric~P Xing, and Masashi
  Sugiyama.
\newblock High-dimensional feature selection by feature-wise kernelized lasso.
\newblock {\em Neural Computation}, 26(1):185--207, 2014.

\bibitem{yamada2018ultra}
Makoto Yamada, Jiliang Tang, Jose Lugo-Martinez, Ermin Hodzic, Raunak Shrestha,
  Avishek Saha, Hua Ouyang, Dawei Yin, Hiroshi Mamitsuka, Cenk Sahinalp, et~al.
\newblock Ultra high-dimensional nonlinear feature selection for big biological
  data.
\newblock {\em IEEE Transactions on Knowledge and Data Engineering},
  30(7):1352--1365, 2018.

\bibitem{10.1093/bioinformatics/btz333}
Héctor Climente-González, Chloé-Agathe Azencott, Samuel Kaski, and Makoto
  Yamada.
\newblock {Block HSIC Lasso: model-free biomarker detection for ultra-high
  dimensional data}.
\newblock {\em Bioinformatics}, 35(14):i427--i435, 07 2019.

\bibitem{efron2004least}
Bradley Efron, Trevor Hastie, Iain Johnstone, Robert Tibshirani, et~al.
\newblock Least angle regression.
\newblock {\em The Annals of statistics}, 32(2):407--499, 2004.

\bibitem{peng2005feature}
Hanchuan Peng, Fuhui Long, and Chris Ding.
\newblock Feature selection based on mutual information: criteria of
  max-dependency, max-relevance, and min-redundancy.
\newblock {\em IEEE Transactions on Pattern Analysis \& Machine Intelligence},
  (8):1226--1238, 2005.

\bibitem{gretton2005measuring}
Arthur Gretton, Olivier Bousquet, Alex Smola, and Bernhard Sch{\"o}lkopf.
\newblock Measuring statistical dependence with hilbert-schmidt norms.
\newblock In {\em International conference on algorithmic learning theory},
  pages 63--77. Springer, 2005.

\bibitem{martens2013explaining}
David Martens and Foster Provost.
\newblock Explaining data-driven document classifications.
\newblock {\em Mis Quarterly}, 38(1):73--100, 2014.

\end{thebibliography}

\end{document}